\documentclass{article}

\usepackage[preprint]{neurips_2026}

\usepackage[utf8]{inputenc} 
\usepackage[T1]{fontenc}    
\usepackage{hyperref}       
\usepackage{url}            
\usepackage{booktabs}       
\usepackage{amsfonts}       
\usepackage{nicefrac}       
\usepackage{microtype}      
\usepackage{xcolor}         

\usepackage{color, colortbl}
\definecolor{LightCyan}{rgb}{0.88,1,1}
\definecolor{LightGreen}{rgb}{0.56, 0.93, 0.56}
\newcommand{\CC}[1]{\cellcolor{LightGreen}}
\usepackage{enumitem}
\usepackage{relsize} 

\usepackage{titletoc}

\usepackage{amssymb}
\usepackage{mathtools}
\usepackage{amsthm}
\usepackage{multirow}
\usepackage{graphicx}
\usepackage{wrapfig}
\usepackage{tablefootnote}

\usepackage{booktabs}
\usepackage{adjustbox}
\usepackage{colortbl}
\usepackage{subcaption}

\usepackage{algorithm}
\usepackage[noend]{algpseudocode}
\usepackage{amsmath} 

\theoremstyle{plain}
\newtheorem{theorem}{Theorem}[section]
\newtheorem{proposition}[theorem]{Proposition}
\newtheorem{lemma}[theorem]{Lemma}
\newtheorem{corollary}[theorem]{Corollary}
\theoremstyle{definition}
\newtheorem{definition}[theorem]{Definition}
\newtheorem{assumption}[theorem]{Assumption}
\theoremstyle{remark}
\newtheorem{remark}[theorem]{Remark}

\usepackage{todonotes}

\usepackage[capitalize,noabbrev]{cleveref}
\crefname{theorem}{Theorem}{Theorems}
\crefname{proposition}{Proposition}{Propositions}
\crefname{lemma}{Lemma}{Lemmas}
\crefname{corollary}{Corollary}{Corollaries}
\crefname{definition}{Definition}{Definitions}
\crefname{assumption}{Assumption}{Assumptions}
\crefname{remark}{Remark}{Remarks}

\title{SMA: Submodular Modality Aligner For Data Efficient Multimodal Learning}

\author{%
  Truong Pham\thanks{Equal Contributions.} \qquad \qquad
  Anay Majee\footnotemark[1] \qquad \qquad
  Rishabh Iyer \\
  The University of Texas at Dallas\\
  \texttt{firstname.lastname@utdallas.edu} \\
}

\begin{document}

\maketitle

\begin{abstract}

Despite the recent success of Multimodal Foundation Models (FMs), their reliance on massive paired datasets limits their applicability in low-data and rare-scenario settings where aligned data is scarce and expensive. A key bottleneck is the adoption of an instance-level formulation, which learns alignment by maximizing correlation between individual image-text pairs while neglecting the underlying geometric structure across modalities resulting in a modality gap across input modalities.
In this paper, we propose a combinatorial paradigm for multimodal alignment that moves beyond pairwise learning and introduce the \emph{Submodular Modality Aligner (SMA)}, which treats multiple augmentations and descriptions of an entity as a set, leveraging multiple descriptions of the data to capture richer cross-modal structure. We instantiate SMA using a principled objective based on Submodular Mutual Information (SMI), which jointly maximizes inter-modality mutual information while reducing cross-modal divergence. This formulation enables the model to effectively utilize multiple positive associations and extract significantly more information from limited data. We evaluate SMA on 14 zero-shot classification and retrieval tasks from the CLIP benchmark and demonstrate consistent gains in the low-data regime. Notably, SMA achieves strong multimodal generalization using only tens of thousands of samples. This is orders of magnitude fewer than standard approaches. Our results highlight the importance of set-based formulations and submodular objectives for data-efficient multimodal learning.\looseness-1
\end{abstract}

\section{Introduction}
\label{sec:intro}
Interacting with the real-world oftentimes involves comprehending a subset of visual, auditory, textual etc. signals together towards decision making tasks. Although unimodal foundational vision and language models have shown commendable improvements in real-world understanding, reasoning and free-form generation tasks, they do not generalize to multimodal tasks. Pioneering works in multimodal learning like CLIP~\cite{DBLP:journals/corr/abs-2103-00020-CLIP}, SigLIP~\cite{Zhai2023SigmoidLF-SigLip} etc. have challenged the status-quo by projecting unimodal feature representations into a shared space to facilitate modeling of cross-modal interactions relevant for downstream understanding and reasoning tasks. \looseness-1

Even though these models achieve universal alignment across modalities, they rely on large scale paired data points (millions if not billions) for learning model alignment which maybe infeasible to collect for rare scenarios. In practice, paired data is frequently \textit{scarce, expensive, or inherently limited}, leading to a regime where only a small number of aligned samples are available and standard contrastive objectives (e.g., InfoNCE) tend to overfit, suffer from poor negative sampling, and fail to generalize. This challenge is particularly prominent in domains such as medical imaging (radiology images paired with expert reports), remote sensing (satellite imagery with limited annotations), autonomous driving (rare edge-case scenarios), security/surveillance (weak or delayed labeling), and scientific/biological data, where collecting large-scale paired datasets is difficult or infeasible.\looseness-1

\begin{figure*}[t]
        \centering
        \includegraphics[width=\textwidth]{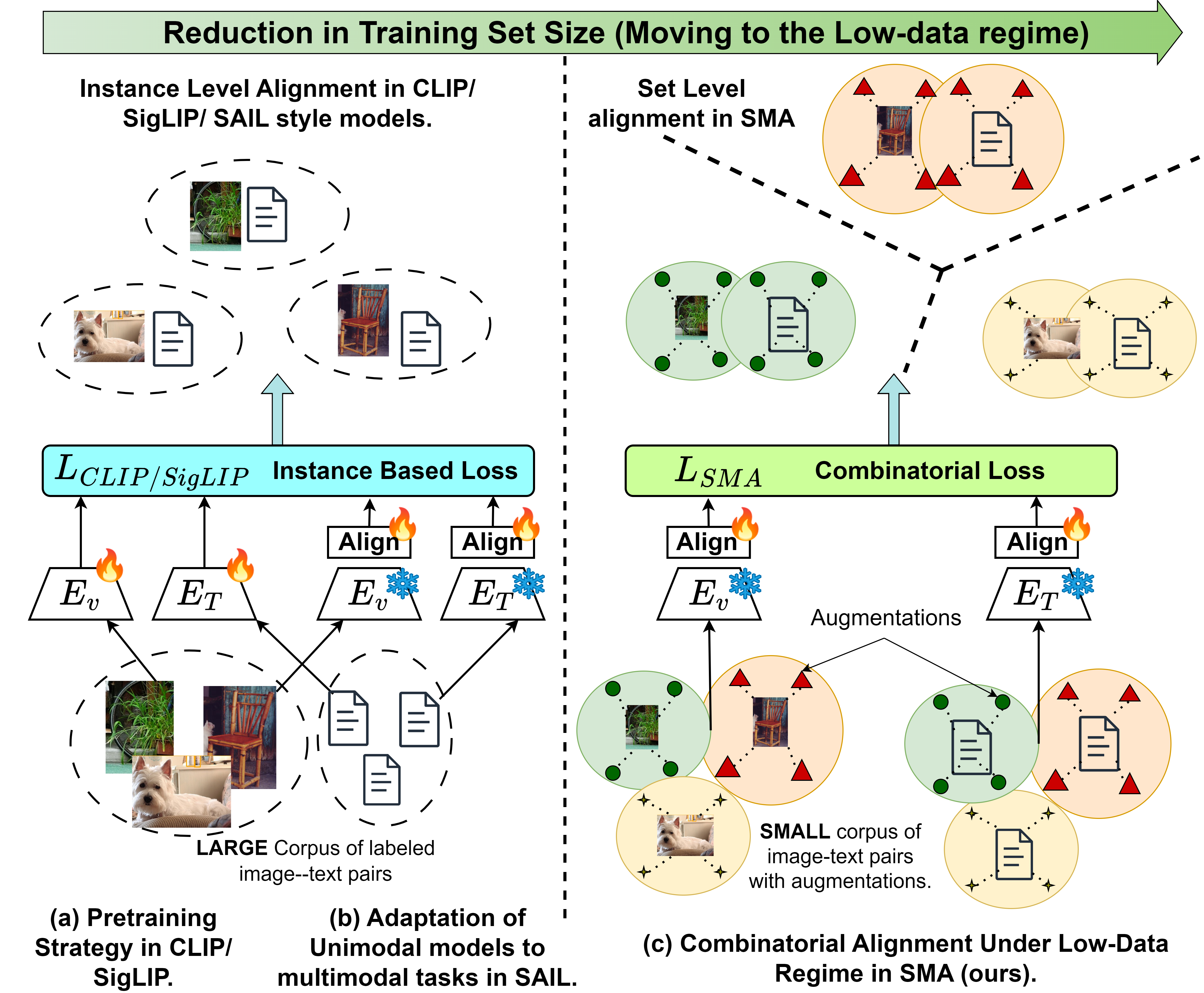}
        \caption{\small \textbf{Architcture of SMA.} Comparison between CLIP, SigLIP, SAIL and our alginment structure. SAIL has frozen pretrained encoders and only trains on a small projection layer, in contrast to the end-to-end training by CLIP and SigLIP. However, all 3 methods are instance based alignment training and can only extract information from singleton positive pairs. Our SMA losses are trained on top of frozen encoders and can efficiently exploit the shared representation information across modalities of a single object.  \looseness-1}
        \label{fig:title_fig}
\end{figure*}

In this work, we introduce a \textit{combinatorial (set-based) formulation into multimodal learning} to adapt unimodal vision and language models to low-data downstream zero-shot tasks under both data and compute constraints.
We build on the observation that present datasets used for multimodal alignment are not necessary combinatorial (instance based) and methods built on them aim at modeling the correlation between a \textit{single} pair of associated image and text embedding of an object without taking into consideration the \textit{multiple} views and descriptions to describe one object. 
\emph{We argue that this limitation is not merely empirical but structural: pairwise contrastive objectives are fundamentally insufficient to model the multi-view nature of real-world entities, leading to inefficient use of data and persistent modality gaps.} In contrast to existing works, in \cref{sec:submod_align_loss} we reformulate existing datasets by arguing that a unimodal datapoint and its multiple descriptions constitute a set, injecting a combinatorial view. Unlike prior extensions of contrastive learning that incorporate multiple positives independently, our formulation explicitly models their joint combinatorial structure as a set.


A key limitation of pairwise contrastive learning is that it treats each positive pair independently, even when multiple views or descriptions correspond to the same underlying entity. This leads to two issues: (i) inefficient use of data, since each pair captures only a fraction of the available semantic information, and (ii) conflicting optimization signals, where different views of the same entity may be pushed apart due to negative sampling. As a result, existing methods fail to fully exploit multi-view structure and struggle to reduce the modality gap, particularly in low-data regimes. Additionally, we observe from \cite{Yin2025DistributionalVA, Liang2022MindTG, shi2023towards} that the robustness of a multimodal alignment model is its ability to jointly maximize \textit{Mutual Information} (MI) and minimize \textit{divergence} across sets from orthogonal modalities, and that the widely observed modality gap is in part a consequence of pairwise contrastive formulations.
Maximizing MI ensures statistical correlation between image and text embeddings. 
On the other hand, minimizing divergence ensures associated image and text datapoints are closer in the embedding space.
While most recent techniques like CLIP~\cite{DBLP:journals/corr/abs-2103-00020-CLIP}, SigLIP~\cite{Zhai2023SigmoidLF-SigLip} and SAIL~\cite{Zhang2024AssessingAL} optimize for alignment, they often overlook the minimization of the modality gap (also shown in \cref{tab:l2distance}).

In a quest to create \textbf{data efficient alignment loss} we introduce a novel learning objective (loss) formulation, namely \textbf{S}ubmodular \textbf{M}odality \textbf{A}ligner (\textbf{SMA}) based on Submodular Mutual Information (SMI). 
As shown in \cref{fig:title_fig} we show that SMI can effectively correlate between multiple positive pairs of datapoints, unlike CLIP \cite{DBLP:journals/corr/abs-2103-00020-CLIP}, SigLIP \cite{Zhai2023SigmoidLF-SigLip} and SAIL \cite{Zhang2024AssessingAL}.
This allows our formulation to rapidly generalize unimodal architectures to multimodal tasks in the low-data regime with only tens of thousands of human annotated examples rather than hundreds of millions of paired exemplars.\looseness-1


We evaluate our model on 14 different tasks with varying difficulties in the CLIP benchmark spanning zero-shot classification and retrieval and demonstrate consistent performance gains in the limited data regime. Our contributions can be summarized as -\looseness-1
\begin{itemize}[leftmargin=*]
    \item We introduce a novel combinatorial, set-based formulation for multimodal alignment that models multiple views and descriptions of an entity jointly rather than independently, enabling richer supervision in low-data regimes.
    
    \item We show that this formulation naturally enables joint modeling of alignment (correlation) and modality gap (divergence), and instantiate it via a principled objective based on Submodular Mutual Information (SMI).
    
    \item We demonstrate that SMA achieves strong empirical gains, with up to 25\% relative improvement and consistent performance across 14 tasks, while using orders of magnitude fewer samples.
\end{itemize}

Beyond empirical gains, our work highlights a broader conceptual shift: multimodal alignment should be formulated over sets of related instances rather than isolated pairs.

\section{Related Work}
\label{sec:rel_work}
\textbf{Multimodal Alignment foundation.} Multimodel alignment methods project the embeddings from unimodal models into a shared space to correlate different representations of the same object, in this case texts and images. The shared space is used for general-purpose downstream tasks such as zero-classification, zero-shot retrieval, question answering, etc. The majority of multimodal alignment methods conform to CLIP \cite{DBLP:journals/corr/abs-2103-00020-CLIP} and ALIGN \cite{Jia2021ScalingUV} approaches by using contrastive learning \cite{Zhai2023SigmoidLF-SigLip, Yuksekgonul2022WhenAW, agarwal2026liteembedadaptingcliprare}. Despite its success, the standard CLIP training paradigm requires training both unimodal encoders and their linear alignment layers from scratch using hundreds of millions of data points, a process that is highly time-consuming and demands prohibitive computational resources \cite{DBLP:journals/corr/abs-2103-00020-CLIP}. Recent work by \cite{Zhang2024AssessingAL} shows that higher performance can be achieved much more efficiently by freezing strong pretrained unimodal encoders such as DinoV2 \cite{Oquab2023DINOv2LR} and NVEmbed \cite{Lee2024NVEmbedIT} and limits training only on the inexpensive alignment layers. Investigations into the aligning power of CLIP's variant of the InfoNCE contrastive loss \cite{DBLP:journals/corr/abs-2103-00020-CLIP} has shown that while correlation between text and image embeddings of the same object can be learned, the projected unimodal embeddings do not align on top of each other \cite{Yin2025DistributionalVA, shi2023towards, Liang2022MindTG, zhou2025clippaeprojectionaugmentationembeddingextract, Liang2022MindTG} demonstrates a phenomenon called the "modality gap", exacerbated by InfoNCE's two conflicting components: the alignment component that pulls embeddings of the same representation together and the uniform embedding distribution that pushes away any embedding not representing the exact same object.\looseness-1

\textbf{Efficient Alignment.} Many works try to find fast and data efficient ways to align unimodal embeddings by leveraging frozen pretrained encoders. ASIF \cite{norelli2023asifcoupleddataturns} proposes a fully training free method by applying nearest neighbor like method onto the extracted embeddings of the training data for retrieval tasks. Other methods like CSA \cite{Li2024CSADM} and CKA \cite{maniparambil2024visionlanguageencodersrepresent} analyze the latent repsentation of unimodal embeddings to find shared representations across modalities. However, these methods do not create a shared embedding space but rely only on the structure of the unimodal spaces and their sample level correlation. Recent methods that create a projection space to unite all modalities rely on CLIP with additional components to extract more information from smaller datasets. \cite{Grger2025WithLD} added regularizations to preserve the neighborhood structures in unimodal embeddings spaces, keeping the geometric semantics of the pretrained models. \cite{vouitsis2024dataefficientmultimodalfusionsingle} is a mixup like method to augments the embeddings for alignment training.\looseness-1

\textbf{Submodularity in Machine Learning.} Combinatorial functions is a niche method in Machine Learning that is gaining foothold in a wide range of applications through Submodular Functions \cite{bilmes2022submodularity}. These functions have been extensively researched in areas such as data subset selection~\cite{killamsetty_automata, prism, jain2023efficient,durga2021training,killamsetty2022nested}, active learning~\cite{wei15_subset, talisman, Beck2021EffectiveEO, Kaushal_2019}, and video summarization~\cite{vid_sum_2019, kaushal2019framework, kaushal2021vid_summary}. Submodular functions also proved to be powerful continuous loss, specifically for contrastive learning \cite{score, smile, majee2025lookingknowndatadiscovery, song2017deepmetriclearningfacility}. Maximizing submodular functions create a diverse representative \cite{JegelkaB2011} set while minimizing them encourage cooperation and representation \cite{kumari2024end}. Both are desirable properties for modeling diverse feature clusters in representation learning tasks. Naturally, Submodular functions are promising candidates from alignment learning, since the first successful and most popular alignment framework is a contrastive learning one \cite{DBLP:journals/corr/abs-2103-00020-CLIP}.\looseness-1

\section{The Submodular Multimodal Aligner Framework}
\label{sec:method} 
\subsection{Preliminaries: Submodularity}
\label{sec:submod_prelims}
Submodular functions~\cite{fujishige,iyer2015polyhedral,bilmes2022submodularity} are set functions popular for being to analytically describe diminishing returns. In particular, a set function $f: 2^\mathcal{V} \rightarrow \mathtt{R}$, defined over a ground set \(\mathcal{V}\), is considered submodular iff it satisfies the inequality $f(A_i) + f(A_j) \geq f(A_i \cup A_j) + f(A_i \cap A_j)$ for all subsets $A_i, A_j \subseteq \mathcal{V}$~\citep{fujishige}. Traditionally, subset selection and summarization tasks are modeled as a discrete optimization problem through submodular maximization~\cite{fujishige, Nemhauser1978} under a knapsack constraint~\cite{Nemhauser1978}. This can be fairly approximated with a $(1 - e^{-1})$ constant factor guarantee~\cite{Nemhauser1978} using greedy optimization techniques~\cite{Nemhauser1978,Mirzasoleiman2015lazierthanlazy}. The \textbf{Submodular Mutual Information Functions} $I_f(A;Q)$ (SMIs) are Information Measure versions of Submodular Functions \cite{iyer2021generalized}, 
\begin{wraptable}{r}{0.52\textwidth}
\centering
\caption{Formulations of FLVMI and FLQMI.}
\label{tab:flvmi_flqmi}
\renewcommand{\arraystretch}{2.0}
\begin{tabular}{lc}
\toprule
\textbf{Method} & \textbf{Formulation} \\
\midrule
FLVMI & $\displaystyle\sum_{i \in U} \min\!\left(\max_{j \in A} s_{ij},\ \max_{j \in Q} s_{ij}\right)$ \\[6pt]
FLQMI & $\displaystyle\sum_{i \in Q} \max_{j \in A} s_{ij} + \sum_{i \in A} \max_{j \in Q} s_{ij}$ \\
\bottomrule
\end{tabular}
\end{wraptable}
specifically the Combinatorial version of Mutual Information. Indeed, Entropic Functions are known to be strictly subsets of Submodular Functions \cite{iyer2021generalized, iyer2026explicitentropicconstructionscoverage}, that is, \textit{all} Entropic Functions are Submodular. Although Mutual Information Function is not entropic, SMIs are submodular if $Q$ is a constant set. Maximizing \textbf{Submodular Information Functions} (SIMs)~\cite{iyer2021submodular,iyer2021generalized,bilmes2022submodularity} $f(A)$ like Facility-Location, Graph-Cut etc. promotes selection of diverse examples within a set $A$, while maximizing \textbf{Submodular Mutual Information Functions} $I_f(A;Q)$ selects examples that share maximum information in $A \cap Q$. Interestingly, recent approaches \cite{score, smile} have also applied combinatorial functions as learning objectives in continuous optimization problems. Specifically, \cite{score} introduces combinatorially inspired loss functions (in this case SIMs) which model intra-group compactness (cooperation)  when minimized and inter-group separation when maximized. Additionally, \cite{smile} introduces SMI as loss function to model interactions between abundant and rare examples in data-efficient (few-shot) representation learning tasks. 

We extend the application of combinatorial loss to align through two SMI functions: Facility Location Submodular Mutual Information Function (FLVMI) and Facility Location Submodular Variant Mutual Infomration Function (FLQMI) \cite{prism} (Table~\ref{tab:flvmi_flqmi}). In targeted subset selection, FLVMI and FLQMI are known for their ability to effectively represent the query set \cite{similar, Beck2024TheoreticalAO}. This means, higher the value of $I_f$, the more representative $A$ is of $Q$. By setting $A$ and $Q$ according to their labels, we can manipulate the embedding space using gradient descent to force $A$ to represent $Q$, which is relevant to our case aligning image and text embeddings.




\subsection{Submodular Multimodal Aligner Objective}
\label{sec:submod_align_loss}
Existing multimodal alignment objectives such as CLIP and SigLIP operate on pairwise similarities between image and text embeddings, differing primarily in how these similarities are aggregated (e.g., softmax-based contrastive normalization or sigmoid-based binary classification). In contrast, we lift this formulation to operate on \emph{sets} of embeddings corresponding to multiple views and descriptions of an entity. This allows us to replace pairwise similarity scores with set-level information measures, leading to a more expressive alignment objective.
We adopt the same training scheme as CLIP for the alignment task, using contrastive loss to align positive pairs together and achieve a uniform distribution of the embeddings \cite{Wang2020UnderstandingCR}. 

Specifically, we partition the data into two parts: the positive sets and the negative sets. A collection of positive pairs, embeddings that represent the same object, will form a positive set $A^{+}_i$, with $i \in \{1,2,..., \text{N}\}$ and N is the number of objects in the dataset. Intuitively, each positive set $A_i^+$ contains multiple views (image augmentations) and descriptions (text variants) of the same underlying entity, while $A_i^-$ captures mismatched image-text combinations across entities. Given a ground set $V=\{\{x_i,y_i\}\}^N_1$ that contains all pairs in the dataset, every negative set $A^{-}_i$ is a negative counterpart of a positive set $A^{+}_i$ with all negative pairs from the positive set, meaning $A^-_i=\{\{x,y\}| x \in A^{+}_i, y \notin A^{+}_i\}\cup \{\{x,y\}|x \notin A^{+}_i, y \in A^{+}_i\}$ . Note that size of $V$ is D, which contains all image and text instances which can be multiple for each object. We further define the partition of the ground set $V$ to be the union of all positive sets $V = \{A^{+}_1 \cup A^{+}_2 \cup ... \cup A^{+}_N\}$, and $A^{+}_1 \cap A^{+}_2 \cap ... \cap A^{+}_N =\emptyset $ or there are no data pairs that represent multiple objects. As with contrastive learning, the embeddings in the positive sets should be pulled together, representing the \textit{alignment} component of the contrastive loss; while the embeddings in the negative sets should be pulled away from their positive counterpart to create a \textit{uniform} representation. We provide a general definition for Submodular Multimodal Alignment loss:

\begin{definition}
    Given a ground set $V = \{x_i, y_i\}^D_1$ comprises of $D$ embedding pairs $x$ and $y$, and a SMI function $I_f$, we provide a Submodular Multimodal Alignment loss:
    \begin{align}
        L_{SMA} = \sum_{i=1}^N I_f(A^{x+}_i,A^{y+}_i) - \sum_{i=1}^N I_f(A^{x-}_i,A^{y-}_i)
    \end{align}
\end{definition}
Here, $V = \{A^{+}_1 \cup A^{+}_2 \cup ... \cup A^{+}_N\}$; $A^{+}_1 \cap A^{+}_2 \cap ... \cap A^{+}_N =\emptyset $; $A^-_i=\{\{x,y\}| x \in A^{+}_i, y \notin A^{+}_i\}\cup \{\{x,y\}|x \notin A^{+}_i, y \in A^{+}_i\}$; $A^x_i = \{x \in A_i\}$  and $A^y_i = \{y \in A_i\}$.

This objective encourages high mutual information between sets of aligned image and text embeddings while simultaneously reducing information shared between mismatched sets.

\begin{figure*}[t]
        \centering
        \includegraphics[width=\textwidth]{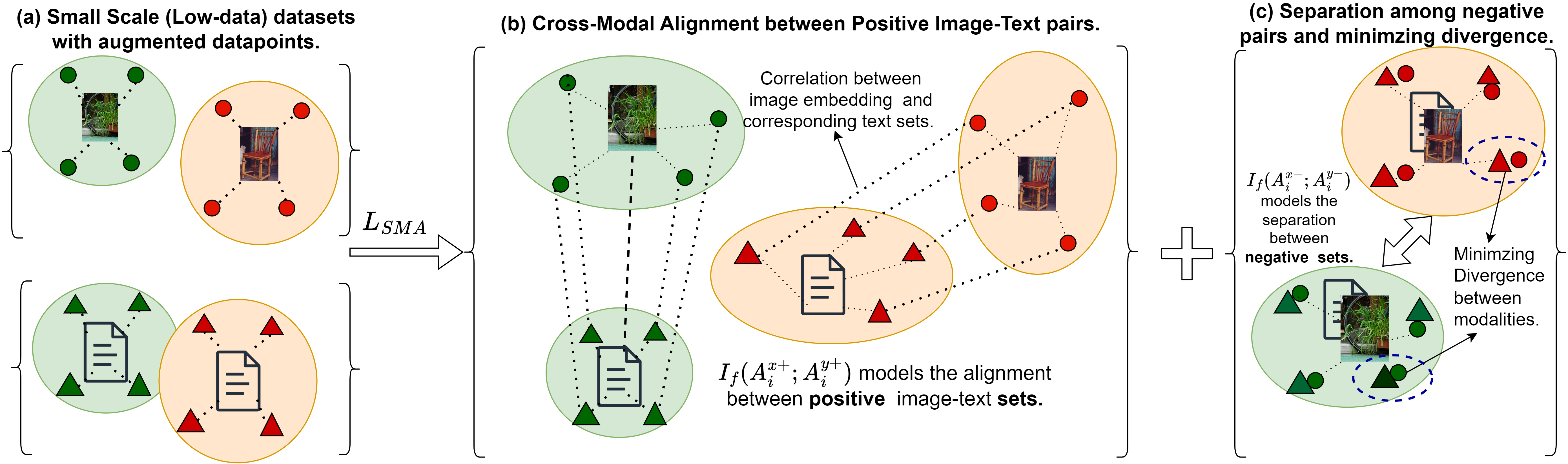}
        \caption{\small \textbf{Illustration of the Loss formulation in Submodular Modality Aligner (SMA)} - Given a image and text pair we first (a) augment them separately alleviating the need for large datasets. Then we train only the alignment layers using the combinatorial SMA loss formulation $L_{SMA}$ which jointly models (b) cross-modal alignments (correlations between image and text sets)  and (c) minimizes divergence across modalities. \looseness-1}
        \label{fig:title_fig}
\end{figure*}

Following the training scheme by CLIP, we also perform alignment from image to text and text to image in the training loop \cite{agarwal2026liteembedadaptingcliprare}. We provide two instances of $L_{SMA}$ as defined earlier by varying the choice of SMI functions for the alignment loss (FLVMI and FLQMI as defined in \cref{sec:submod_prelims}): 
\begin{definition}
    Facility Location Information Alignment loss or FLVMIA loss:
    \begin{align}
        L_{SMA}^{FLVMI} = \sum_{i=1}^N\sum_{j\in A^{x+}_i} \min \left( \max_{k\in A^{x+}_i} s_{jk},\max_{k\in A^{y+}_i} s_{jk}\right) - \sum_{i=1}^N\sum_{j\in A^{x-}_i} \min\left(\max_{k\in A^{x-}_i} s_{jk}, \max_{k\in A^{y-}_i} s_{jk}\right)
    \end{align}
\end{definition}
Note, that computation of $I_f$ requires modeling of interactions between datapoints which we achieve through the cosine similarity kernel as used in combinatorial representation learning literature\cite{score,smile}. 
Defined in \cite{prism}, FLVMI is widely used in guided subset selection to select a subset most representative of its query set \cite{killamsetty_automata} (add more subset selection papers). In this case, the query set is replaced by half of the positive pair or negative pairs and the selected subset is already chosen to be the other half. Maximizing this function will force $A^{x+}$ to be representative of $A^{y+}$ and pushes $A^{x-}$ away from $A^{y-}$. This is similar to how contrastive loss is being used in representation learning, where embeddings of similar classes should cluster together and move away from the other classes \cite{score}. We also introduce a variant of FLVMI:

\begin{definition}
    Facility Location Variant Mutual Information Alignment loss or FLQMIA loss:\\
    \begin{align}
        L_{SMA}^{FLQMI} = \sum_{i=1}^N\sum_{j\in A^{x+}_i} \max_{k\in A^{y+}_i} s_{jk} - \sum_{i=1}^N\sum_{j\in A^{x-}_i} \max_{k\in A^{y-}_i} s_{jk}
    \end{align}
\end{definition}

Intuitively, these objectives encourage each element in one modality (e.g., image embeddings) to be well-represented by at least one element in the corresponding set from the other modality (e.g., text embeddings), capturing cross-modal coverage and alignment. FLQMI is a different form of FLVMI \cite{prism}. While both belong to the family of facility-location-based submodular mutual information functions, they differ in how cross-set interactions are aggregated. In particular, FLVMI emphasizes coverage through a min-max structure, encouraging balanced representation across modalities, whereas FLQMI adopts a max-based formulation that focuses on strongest cross-modal associations.

\subsection{Connections to CLIP and SigLIP}
We now show how our formulation relates to widely used multimodal alignment objectives such as CLIP and SigLIP. These methods operate on pairwise similarities between image and text embeddings, differing primarily in how these similarities are aggregated across positive and negative pairs.

\noindent \textbf{Reduction to CLIP.}
In the special case where each object is associated with a single image-text pair, the positive sets reduce to singletons, i.e., $|A_i^+| = 1$. Under this setting, the FLQMIA objective simplifies significantly. Using the log-sum-exp approximation to the max operator, we obtain:
\begin{align*}
\sum_{i=1}^N \log\left(e^{s_{ii}}\right) - \sum_{i=1}^N \log\left(\sum_{j \in A_i^-} e^{s_{ij}}\right)
= \sum_{i=1}^N \log \left( \frac{e^{s_{ii}}}{\sum_{j \in A_i^-} e^{s_{ij}}} \right).
\end{align*}

This is precisely the NT-Xent loss used in SimCLR \cite{Chen2020ASF}, and closely matches the CLIP objective \cite{DBLP:journals/corr/abs-2103-00020-CLIP}, showing that CLIP-style contrastive learning arises as a special case of our formulation under singleton set constructions.

\noindent \textbf{Relation to SigLIP.}
SigLIP adopts a different pairwise alignment strategy based on independent binary classification over image-text pairs using a sigmoid loss. Specifically, it optimizes a softplus-based objective of the form $\log(1 + \exp(-z_{ij}s_{ij}))$, where $z_{ij}$ denotes whether a pair is matched or mismatched. This can be interpreted as a smooth max-margin objective, where each pair is treated independently. In contrast, our formulation replaces pairwise similarity scores with set-level submodular mutual information. While SigLIP operates on individual pairs, our framework jointly models interactions across multiple views and descriptions within each set. Thus, SigLIP can be viewed as a pairwise member of the broader family of alignment objectives that our framework lifts to the set level.

\noindent Overall, our formulation generalizes multimodal alignment objectives from pairwise similarity functions to set-level information measures. Standard objectives such as CLIP and SigLIP emerge as special or degenerate cases when sets collapse to singletons, whereas our formulation enables richer modeling of multi-view structure in the general case.

\subsection{Connection to the Modality Gap}
CLIP and its variance have been shown to suffer from Modality Gap. This is a phenomenon where a linear separation can be found between the image and text embeddings even in the shared projection space \cite{shi2023towards}. In brief details, the two components of InfoNCE: alignment and uniformity, are necessary but opposing forces, requiring a balancing between complete alingment which pulls embeddings together and uniformity which pushes embeddings away. We shows that solving the modality gap is a submodular problem (Appendix \ref{appendix:MG-Scon}).
\begin{remark}
    Minimizing the Modality Gap is the same as maximizing the Submodular Mutual Information Function
\end{remark}
The proof from Appendix \ref{appendix:MG-Scon} shows that maximizing the SMI function $I_f(A^x,A^y)$ where $A=V$, meaning the subset $A$ covers everything, will effectively reduce the Modality Gap. This is a direct link between the Modality Gap and Submodular Mutual Information function. However, this connection ignores the other crucial component alignment contrastive loss: the uniformity or the negative component. This leads to a conclusion that while SMI can help alleviate the modality gap phenomenon, it is not the cure to this problem. This is supported by our experimental results that shows smaller modality gaps between distributions when using SMA but not an entire overlap (Table~\ref{tab:l2distance}).\looseness-1

\section{Experiments}
\label{sec:exp}

\paragraph{Datasets} - Since the primary goal of this work is to learn under data scarcity, we use the MS-COCO~\cite{coco} Captions dataset as the primary training data to extract maximal cross-modal information from limited paired data rather than relying on large-scale pretraining. Each image in MS-COCO is associated with at least 5 different captions (human annotated) which serve as natural augmentations while being a relatively small-scale benchmark of 80K images. 
This also follows prior works in data-efficient multimodal alignment such as STRUCTURE~\cite{Grger2025WithLD} and CSA~\cite{Li2024CSADM} who also adopt MS-COCO (or its subset) during alignment training.
We evaluate on 14 zero shot classification tasks and Flickr30k for zero shot retrieval task using the Clip Benchmark test~\cite{DBLP:journals/corr/abs-2103-00020-CLIP}.\looseness-1

\begin{table}[t]
    \centering
    \caption{\textbf{Benchmark Results on Zero-shot classification} accuracy (\%) across benchmarks for multiple positive pair setting. Alignment is performed on frozen pretrained DINOv2 and GTE-1.5. Best results are in \colorbox{orange!60}{\textbf{Bold}} while the second best in in \colorbox{orange!20}{\textit{italics}}.}
    \vspace{0.5em}
    \begin{subtable}{\textwidth}
        \centering
        \adjustbox{max width=\textwidth}{
        \begin{tabular}{l *{7}{c}}
            \toprule
            \textbf{Method} & \textbf{Caltech101} & \textbf{Cars} & \textbf{CIFAR-10} & \textbf{DTD} & \textbf{EuroSAT} & \textbf{FGVC} & \textbf{Flowers} \\
            \midrule
            SAIL w/ CLIP \cite{Zhang2024AssessingAL}& 55.66 & \cellcolor{orange!20}\textit{2.55} & 92.32 & 13.99 & 34.70 & \cellcolor{orange!20}\textit{2.37} & 7.17 \\
            SAIL w/ SigLIP \cite{Zhang2024AssessingAL}      & \cellcolor{orange!20}\textit{58.23} & \cellcolor{orange!20}\textit{2.55} & 93.16 & 17.18 & 32.84 & 1.74 & \cellcolor{orange!60}\textbf{8.75} \\
            CSA \cite{Li2024CSADM}         & 46.28 & 1.42 & 87.19 & 17.29 & 26.00 & \cellcolor{orange!60}\textbf{3.51} & 5.24 \\ \midrule
            FLQMIA       & \cellcolor{orange!60}\textbf{59.06} & \cellcolor{orange!60}\textbf{3.15} & \cellcolor{orange!60}\textbf{94.42} & \cellcolor{orange!60}\textbf{21.44} & \cellcolor{orange!60}\textbf{40.18} & 1.83 & \cellcolor{orange!20}\textit{8.28} \\
            FLVMIA       & 53.30 & 2.29 & \textit{93.54} & \cellcolor{orange!20}\textit{21.60} & \cellcolor{orange!20}\textit{39.93} & \cellcolor{orange!20}\textit{2.37} & 6.57 \\
            \bottomrule
        \end{tabular}
        }
    \end{subtable}

    \vspace{0.5em}

    \begin{subtable}{\textwidth}
        \centering
        \adjustbox{max width=\textwidth}{
        \begin{tabular}{l *{7}{c}}
            \toprule
            \textbf{Method} & \textbf{GTSRB} & \textbf{Pets} & \textbf{STL-10} & \textbf{Imagenet-v1} & \textbf{Food101} & \textbf{CIFAR-100} & \textbf{MMVP} \\
            \midrule
            SAIL w/ CLIP   \cite{Zhang2024AssessingAL}& 6.48 & 14.06 & 87.83 & 20.20 & 29.61 & 45.22 & 16.30 \\
            SAIL w/ SigLIP \cite{Zhang2024AssessingAL}& 7.20 & 14.58 & 90.62 & 25.03 & 35.28 & 50.50 & 14.81 \\
            CSA \cite{Li2024CSADM}          & 5.78 & 12.56 & 85.40 & 24.62 & 31.76 & 44.18 & 15.56 \\ \midrule
            FLQMIA       & \cellcolor{orange!60}\textbf{8.90} & \cellcolor{orange!20}\textit{15.10} & \cellcolor{orange!60}\textbf{92.65} & \cellcolor{orange!60}\textbf{28.04} & \cellcolor{orange!60}\textbf{39.81} & \cellcolor{orange!60}\textbf{53.31} & \cellcolor{orange!20}\textit{17.78} \\
            FLVMIA       & \textit{7.90} & \cellcolor{orange!60}\textbf{15.97} & \cellcolor{orange!20}\textit{91.34} & \cellcolor{orange!20}\textit{27.12} & \cellcolor{orange!20}\textit{36.68} & \cellcolor{orange!20}\textit{51.69} & \cellcolor{orange!60}\textbf{18.52} \\
            \bottomrule
            
        \end{tabular}
        }
    \end{subtable}
    \label{tab:400kzsc}
\end{table}

\paragraph{Experimental Setup} - We follow the experimental setup of SAIL~\cite{Zhang2024AssessingAL} since it provides a testbed for experimenting with several unimodal models for multimodal adaptation allowing us to evaluate several zero-shot downstream tasks.
We compare SMA to two different types of data efficient alignment methods: CSA which is an analytical method using Cannonical Correlation Analysis~\cite{Li2024CSADM}, and SAIL~\cite{Zhang2024AssessingAL} which modifies the alignment layer to be more efficient with frozen pretrained modal. While SAIL is not made for efficient data training, it follows the CLIP alignment framework but with Gated Linear Unit (GLU) as the alignment layer and SigLIP loss instead of InfoNCE. We also include a version of SAIL with InfoNCE as the alignment loss instead of just SigLIP. For CSA, we use 500 for the latent space dimension. We also use SAIL's GLU for the alignment architecture for fair comparison between the losses. Although SAIL trains on dataset several hundred times bigger than out training dataset, they still use much smaller datasets than CLIP's original dataset with better performance.
To maintain uniformity across tasks we use DINOv2 as the image encoder and GTE-1.5 as the text coder (accounting for compute limitations). For CLIP and SigLIP we maintain a batch size of $\sim$32K (as used in the original paper) while for our SMA objectives we maintain a batch size of $\sim$16K for FLQMIA and $\sim$1K for FLVMIA (computes multiple similarity kernels). We set an initial learning rate of 1e-5 and adopt the Lion~\cite{lion_optim} optimizer for training. The entire experiment is done on one RTX A6000 GPU with 48 GB of memory.\looseness-1

\textbf{Zero-shot classification.} We experiment on an extensive zero-shot image classification baselines with 14 tasks of various difficulties in Table~\ref{tab:400kzsc}. Since we implement our loss functions without the memory optimization from deep learning packages on the massive cross similarity losses, our version of the loss can run only run on 16k batch size for FLQMIA and 1k batch size for FLVMIA, while SAIL is being trained on $\sim$32k batch size. We still train till 50 epoch just like in SAIL, but we also use validation loss to stop overfitting in our loss. Early stopping is also done on SAIL but it's either not needed or the saturation point of validation loss is very close to epoch 50, therefore, this does not effect SAIL's performance. According to Table~\ref{tab:400kzsc}, we see significant gains compare to both versions of SAIL and CSA when using our set formulations on multiple positive pairs dataset. On CIFAR10~\cite{cifar} and STL10~\cite{stl10}, accuracies are close to 90\%, which means they are close to being solved, the gains present but minimal with only 1-2\%. On the other hand, we can expect gains from 15-20\% performance boost percentage wise between our best loss and SAIL's best configuration and CSA. This is true across harder datasets with accuracies close to 5\% and datasets with average accuracies higher than 10\%. The only exceptions are Caltech 101~\cite{caltech101} with low performance gains and Flowers~\cite{flowers} and FGVC~\cite{fgvc} which we do not beat SAIL and CSA. Moreover, \cref{tab:400kzsc} shows both of our set formulations are almost always the second best performing loss if not the best.
\begin{wraptable}{r}{0.38\textwidth}
    \centering
    \small
    \vspace{-1em}
    \caption{Flickr30k zero-shot retrieval Recall@1 (\%).}

    \small
    \setlength{\tabcolsep}{4pt}  
    \textbf{(a) Multiple positive pair}\\[0.3em]
    \begin{tabular}{@{}l cc@{}}
        \toprule
        \textbf{Method} & \textbf{I2T} & \textbf{T2I} \\
        \midrule
        SAIL w/ CLIP \cite{Zhang2024AssessingAL}  & \textit{48.40} &  \textbf{39.16} \\
        SAIL w/ SigLIP \cite{Zhang2024AssessingAL}& 33.70 & 24.90 \\
        CSA  \cite{Li2024CSADM}  & 34.80 & 25.44 \\
        FLVMIA   & \textbf{50.50} & \textit{37.70} \\
        \bottomrule
    \end{tabular}

    \vspace{0.8em}
    \textbf{(b) Single positive pairs}\\[0.3em]
    \begin{tabular}{@{}l cc@{}}
        \toprule
        \textbf{Method} & \textbf{I2T} & \textbf{T2I} \\
        \midrule
        SAIL w/ CLIP \cite{Zhang2024AssessingAL}  & 53.80 & 43.16 \\
        SAIL w/ SigLIP \cite{Zhang2024AssessingAL} & 57.70 & \textit{47.28} \\
        CSA w \cite{Li2024CSADM}   & 44.00 & 30.92 \\
        FLQMIA  & \textbf{63.00} & \textbf{51.24} \\
        FLVMIA   & \textit{59.50} & 47.08 \\
        \bottomrule
    \end{tabular}
    \label{tab:flickr30k}
    \vspace{-2.5em}
\end{wraptable}
This is a good show case of the benefits of alignment set formulations in efficient data regime. by leveraging multiple labels, we can extract more invariant semantics from a smaller training dataset. This performance can be improved in the future with better memory optimization to increase our batch size, helping to include more positive pairs per batch.\looseness-1 
\begin{table}[t]
    \centering
    \caption{\textbf{Results from Zero-shot classification} accuracy (\%) across benchmarks for singular positive pair setting. Alignment is performed on frozen pretrained DINOv2 and GTE-1.5. Best results are in \colorbox{orange!60}{\textbf{Bold}} while the second best in in \colorbox{orange!20}{\textit{italics}}.}
    \vspace{0.5em}

    \begin{subtable}{\textwidth}
        \centering
        \adjustbox{max width=\textwidth}{
        \begin{tabular}{l *{7}{c}}
            \toprule
            \textbf{Method} & \textbf{Caltech101} & \textbf{Cars} & \textbf{CIFAR-10} & \textbf{DTD} & \textbf{EuroSAT} & \textbf{FGVC} & \textbf{Flowers} \\
            \midrule
            SAIL w/ CLIP \cite{Zhang2024AssessingAL}  & \cellcolor{orange!60}\textbf{53.73} & \cellcolor{orange!60}\textbf{2.29} & \cellcolor{orange!20}\textit{91.10} & \cellcolor{orange!60}\textbf{16.12} & \cellcolor{orange!20}\textit{36.30} & 1.56 & \cellcolor{orange!20}\textit{7.40} \\
            SAIL w/ SigLIP \cite{Zhang2024AssessingAL}& 45.59 & 1.42 & 90.30 & 9.84 & 29.99 & \cellcolor{orange!20}\textit{2.16} & 4.10 \\
            CSA  \cite{Li2024CSADM}   & 41.09 & 0.93 & 85.39 & 13.62 & 23.90 & \cellcolor{orange!60}\textbf{2.76} & 4.11 \\
            FLVMIA   & \cellcolor{orange!20}\textit{53.72} & \cellcolor{orange!20}\textit{2.00} & \cellcolor{orange!60}\textbf{92.22} & \cellcolor{orange!20}\textit{14.73} & \cellcolor{orange!60}\textbf{36.33} & 1.74 & \cellcolor{orange!60}\textbf{8.08} \\
            \bottomrule
        \end{tabular}
        }
    \end{subtable}

    \vspace{0.5em}

    \begin{subtable}{\textwidth}
        \centering
        \adjustbox{max width=\textwidth}{
        \begin{tabular}{l *{7}{c}}
            \toprule
            \textbf{Method} & \textbf{GTSRB} & \textbf{Pets} & \textbf{STL-10} & \textbf{Imagenet-v1} & \textbf{Food101} & \textbf{CIFAR-100} & \textbf{MMVP} \\
            \midrule
            SAIL w/ CLIP \cite{Zhang2024AssessingAL}  & 6.75 & \cellcolor{orange!20}\textit{11.47} & \cellcolor{orange!20}\textit{85.38} & 18.90 & 27.48 & \cellcolor{orange!20}\textit{43.54} & 14.07 \\
            SAIL w/ SigLIP \cite{Zhang2024AssessingAL} & 4.36 & \cellcolor{orange!60}\textbf{11.61} & 80.45 & 14.32 & 23.40 & 37.49 & 14.07 \\
            CSA  \cite{Li2024CSADM}  & \cellcolor{orange!20}\textit{7.78} & 8.45 & 83.73 & \cellcolor{orange!20}\textit{19.73} & \cellcolor{orange!20}\textit{29.82} & 40.80 & \cellcolor{orange!60}\textbf{18.52} \\
            FLVMIA   & \cellcolor{orange!60}\textbf{7.92} & 11.23 & \cellcolor{orange!60}\textbf{86.92} & \cellcolor{orange!60}\textbf{21.59} & \cellcolor{orange!60}\textbf{30.86} & \cellcolor{orange!60}\textbf{46.25} & \cellcolor{orange!20}\textit{14.81} \\
            \bottomrule
        \end{tabular}
        }
    \end{subtable}
    \label{tab:80kzsc}
\end{table}

We also include experimental results for single positive pair setting with 80k pairs from MSCOCO dataset. We compare SAIL and CSA with FLVMIA, since FLQMIA is too similar to InfoNCE in this setting and the different batch sizes can affect the comparison. Although this problem affects FLVMIA, since our batch size is only 1k, we are already at a disadvantage. However, Table~\ref{tab:80kzsc} shows FLVMIA only performs worse than SAIL on 5/14 datasets while still being the second best or having minimal difference in accuracy to the better losses. FLVMIA is better than CSA in all but one dataset. Furthermore, we let SAIL trains for longer epochs, with the InfoNCE version of SAIL reaching 200 epochs and original SAIL with SigLIP reaches 92 epochs, while FLVMIA converges at 11 epochs. These increase in epochs compare to the standard 50 epoch by SAIL is because we see those losses can still benefit from more training based on the evaluation loss.

\textbf{Zero-shot Retrieval.} Flickr30k is a standard zero-shot retrieval benchmark for multimodal alignment. We perform the same set of experiments on Flickr30k and yield the same results: FLVMI and FLQMI are better than SAIL and CSA in both single positive pair and multiple positive pair settings. In more detail, FLQMIA performs best with a 4--5\% increase in retrieval accuracy, with FLVMIA being the second best or only a few points off. The only exception is in the single positive pair setting (Table~\ref{tab:flickr30k}) where SAIL with InfoNCE manages to beat our method on the Text-to-Image task. This is consistent with the pattern from zero-shot classification: when there are no multiple positive pairs in the training data, our loss is not as competitive. The strong results in the multiple positive pair setting (Table~\ref{tab:flickr30k}) further evidence our effectiveness at processing multiple positive pairs and their importance in low-data settings such as MSCOCO.

\begin{wraptable}{r}{0.30\textwidth}
    \centering
    \vspace{-1em}
    \caption{avg.\ $\ell_2$ distance ($\downarrow$).}

    \small
    \textbf{(a) Multiple positive pair}\\[0.3em]
    \begin{tabular}{l c}
        \toprule
        \textbf{Method} & \textbf{L2} \\
        \midrule
        SAIL w/ CLIP \cite{Zhang2024AssessingAL}  & 22.28 \\
        SAIL w/ SigLIP \cite{Zhang2024AssessingAL} & 15.40 \\
        FLQMIA  & \textit{13.85} \\
        FLVMIA   & \textbf{12.82} \\
        \bottomrule
    \end{tabular}

    \vspace{0.8em}
    \textbf{(b) Single positive pair}\\[0.3em]
    \begin{tabular}{l c}
        \toprule
        \textbf{Method} & \textbf{L2} \\
        \midrule
        SAIL w/ CLIP  \cite{Zhang2024AssessingAL} & 18.76 \\
        SAIL w/ SigLIP \cite{Zhang2024AssessingAL} & \textit{11.93} \\
        FLVMIA   & \textbf{8.58} \\
        \bottomrule
    \end{tabular}
    \label{tab:l2distance}
    \vspace{-0.5em}
\end{wraptable}

\textbf{Analysis on Modality Gap} We address the claim that our functions are more effective in reducing the alignment gap during training by comparing the average $\ell_2$-norm. In order of least effective to most effective, we have the following: FLQMIA (the multiple positive pair setting), FLVMIA, SAIL, SAIL with InfoNCE. InfoNCE is by far the least effective method to reduce the gap, although this does not necessary reflect the loss' ability to generalize. SigLIP is surprisingly more effective at closing the gap with a 30\% reduction in average distance in the multiple positive pair setting and 36\% in single pair setting. This is a much better reduction than FLVMIA compare to SigLIP which is at 10\% in multiple pair setting and 28\% in single pair setting, while FLQMI is only marginally better than FLVMIA in the multiple pair setting. However, this is still proof that our set formations can both correlate and align modalities better than the popular frameworks.\looseness-1

Overall, we see significant performance gain in the multiple positive pair setting across both tasks (Table~\ref{tab:400kzsc}, Table~\ref{tab:flickr30k}). We see up to 25\% accuracy gain when using multiple positive pairs and an average of 7.7\% proportional gain across zero shot classification tasks (Table~\ref{tab:400kzsc}) and around 5\% again for retrieval tasks (Table~\ref{tab:flickr30k}). These are significant gains across 16 total tasks, giving credence to our set formulations. The gains are weaker when there is only one positive pair per object. However, we are only performing worse in the edge cases where the datasets are very difficult with accuracies close to random noise such as cars and air craft with 192 and 100 classes respectively (Table~\ref{tab:80kzsc}). We still see gains up to 9\% on 66\% of the tasks, despite the fact that we are using much smaller batch size. We are also better than traditional contrastive losses at reducing the gap between modalities (Table~\ref{tab:l2distance}).\looseness-1

\section{Conclusion}
\label{sec:conclusion} 
In this work, we tackle multimodal alignment from a data-efficient lens and argue that current paradigms, despite their success, remain fundamentally limited by their reliance on large-scale paired datasets and instance-level formulations. To address these challenges, we introduced a combinatorial, set-based perspective for multimodal learning and proposed Submodular Modality Aligner (SMA), a novel objective grounded in Submodular Mutual Information. By jointly maximizing mutual information between multiple possitive pairs and minimizing cross-modal divergence, SMA explicitly models both alignment and modality gap two aspects that are often decoupled or under-emphasized in prior approaches.
We demonstrate strong multimodal generalization in the low-data regime, requiring only tens of thousands of samples while consistently outperforming existing baselines across diverse zero-shot classification and retrieval tasks. Beyond performance gains, our formulation highlights the importance of set structure and submodular objectives in learning robust cross-modal representations. We believe this work opens up a promising direction for principled, data-efficient multimodal learning, with future avenues including extending the framework to more modalities (e.g., audio, video), exploring adaptive set construction strategies, and integrating SMA into larger-scale foundation models for broader real-world deployment.


\bibliographystyle{plain}
\bibliography{references}


\appendix
\section{Appendix}

\subsection{Modularity Gap connection to Submodularity}
\label{appendix:MG-Scon}
Consider the Submodular function $f(X) = -(\sum_{x\in X} x)^2$, we have a version of SMI function:\\
\begin{align}
    I_f(X,Y)=& f(X) + f(Y) - f(X\cup Y) \nonumber \\
    =&-\left(\sum_{x\in X}x\right)^2 - \left(\sum_{y\in Y}y\right)^2 + \sum_{x\in X\cup Y}x\sum_{y\in X\cup Y}y \label{eq1}
\end{align}
The Modality Gap is defined as:\\
\begin{definition}
    The Modality Gap is the $\ell_2$ distance between the center mean across modalities \cite{Liang2022MindTG}:
    \begin{align*}
        \left\lVert \frac{\sum_{i\in X} x_i}{D} - \frac{\sum_{j\in Y}y_j}{D}\right\rVert^2
    \end{align*}
\end{definition}
Without loss of generality, the gap can be calculated as:
\begin{align}
\left\lVert \frac{\sum_{i\in X} x_i}{D} - \frac{\sum_{j\in Y}y_j}{D}\right\rVert^2 =&\left(\frac{\sum_{i\in X}x_i}{D}\right)^2 + \left(\frac{\sum_{i\in Y}y_j}{D}\right)^2 - 2\left(\frac{\sum_{i\in X}x_i\sum_{i\in Y}y_j}{D^2}\right) \nonumber \\
    =&\left(\frac{\sum_{i\in X}x_i}{D}\right)^2+ \left(\frac{\sum_{i\in Y}y_j}{D}\right)^2-2\left(\frac{\sum_{i\in X\cup Y}x_i\sum_{j\in X\cup Y}y_j}{D^2}\right) \nonumber \\ 
    &+2\left(\frac{\sum_{i\in X}x_i}{D^2}\right)^2+ 2\left(\frac{\sum_{i\in Y}y_j}{D^2}\right)^2 \nonumber \\
    =&(D+2)\left(\frac{\sum_{i\in X}x_i}{D^2}\right)^2+ (D+2)\left(\frac{\sum_{i\in Y}y_j}{D^2}\right)^2 \label{eq2} \\ 
    &-2\left(\frac{\sum_{i\in X\cup Y}x_i\sum_{j\in X\cup Y}y_j}{D^2}\right) \nonumber
\end{align}

Since $D$ is constant, minimizing Equation~\ref{eq2} is the same as maximizing Equation~\ref{eq1}. Therefore, minimizing the Modality Gap is the same as maximizing Mutual Information Function
\subsection{Social Impact}
This paper presents work whose goal is to advance the field of Machine Learning in general and Representation Learning in particular. There are many potential societal consequences of our work, none which we feel must be specifically highlighted here.

\subsection{Limitations}
\label{sec:limits} 
One major limitation of our method is the unoptimized cross similarity matrices. This is especially apparent in FLVMI since it requires 3 different cross similarity matrices. This problem is less severe in FLQMI which contributed to its better performance compared to FLVMI. Better memory scheme needs to be applied to increase the number of positive pairs and objects per training batch. 

Another limitation is our neglect our other SMIs:  Graph Cut Mutual Information (GCMI) and Log-determinant Mutual Information (LOGDETMI). These are prominent SMIs with applications as contrastive losses in other works \cite{smile, majee2025lookingknowndatadiscovery, score, prism}. However, the balance between positive and negative makes it difficult for GCMI since the negative pairs are multiples of positive pairs and thus disproportionally affect the loss. LOGDETMI is harder to optimize and usually yields worse results than its FL counterparts \cite{score}

\subsection{Evaluation Datasets}
\begin{table*}[h]
\centering
\caption{Datasets used for classification and retrieval tasks.}
\label{tab:datasets}
\begin{tabular}{l l c c c}
\hline
\textbf{Task} & \textbf{Dataset} & \textbf{Number of Classes} & \textbf{Train size} & \textbf{Test size} \\
\hline
\multirow{13}{*}{Classification}
& Caltech101      & 101  & 3,060      & 6,084 \\
& Cars            & 196  & 8,144      & 8,041 \\
& CIFAR-10        & 10   & 50,000     & 10,000 \\
& DTD             & 47   & 3,760      & 1,880 \\
& EuroSAT         & 10   & 10,000     & 5,000 \\
& FGVC Aircraft   & 100  & 6,667      & 3,333 \\
& Flowers         & 102  & 2,040      & 6,149 \\
& GTSRB           & 43   & 26,640     & 12,630 \\
& Pets            & 37   & 3,680      & 3,669 \\
& STL-10          & 10   & 5,000      & 8,000 \\
& Imagenet-v1     & 1000 & 1,281,167  & 50,000 \\
& Food101         & 101  & 75,750     & 25,250 \\
& CIFAR-100        & 100  & 50,000     & 10,000 \\
& MMVP            & 9  & N/A     & 270 \\
\hline
\multirow{1}{*}{Retrieval}
& Flickr30k       & N/A  & 29,783     & 5,000 \\
\hline
\end{tabular}
\end{table*}

\subsection{Training Configurations}
\begin{table}[h]
\centering
\caption{Default hyperparameters and datasets used throughout the paper.}
\label{tab:default_hparams_datasets}
\begin{tabular}{p{3.1cm} p{6.4cm} p{2.2cm}}
\hline
\textbf{Category} & \textbf{Hyperparameter / Dataset} & \textbf{Value} \\
\hline

\multirow{2}{*}{Training setup}
& Training dataset & MS COCO Captions \\
\hline

\multirow{4}{*}{Encoders}
& Image encoder & DINOv2 \\
& Text encoder & GTE-1.5 \\
& Alignment layer & GLU \\
& Frozen backbone & Yes \\
\hline

\multirow{6}{*}{Optimization}
& Optimizer & LION \\
& Initial learning rate & 1e-5 \\
& Batch size (FLQMIA) & 16K \\
& Batch size (FLVMIA) & 1K \\
& Batch size (SAIL baseline) & 32K \\
& Max training epochs - multiple postive pairs pairs & 50 \\
& Max training epochs - singular postive pairs pairs & 200 \\
\hline
\end{tabular}
\end{table}


\end{document}